\documentclass[11pt]{article}

\usepackage[utf8]{inputenc}
\usepackage[T1]{fontenc}
\usepackage{lmodern}
\usepackage[margin=1in]{geometry}
\usepackage{microtype}
\usepackage{booktabs}
\usepackage{tabularx}
\usepackage{amsmath}
\usepackage{amssymb}
\usepackage{textcomp}
\usepackage{enumitem}
\usepackage[hidelinks]{hyperref}
\usepackage[numbers,sort&compress]{natbib}
\usepackage{titlesec}

\newcommand{\chisq}{\ensuremath{\chi^2}}

\titleformat{\section}{\normalfont\large\bfseries}{\thesection}{0.6em}{}
\titleformat{\subsection}{\normalfont\normalsize\bfseries}{\thesubsection}{0.6em}{}

\title{\textbf{Computational Orientalism: Measuring Structural Discourse Bias in Large Language Models Using the Middle East Cultural Sensitivity Score (MECSS)}}

\author{Maha Shahid\\
\small Independent Researcher}

\date{}

\begin{document}

\maketitle

\begin{abstract}
\noindent
AI systems now shape how hundreds of millions of people learn about cultures other than their own. When someone asks one of these systems about the Middle East, they do not receive neutral facts. They receive a representation shaped by the assumptions and frameworks embedded in training data, and that data is overwhelmingly Western and English-language. This paper asks whether that representation is Orientalist in Said's sense: whether it denies agency to Middle Eastern actors, treats Western analytical frameworks as neutral while marking non-Western knowledge as particular, and explains the region through categories it did not produce. Standard fairness metrics cannot answer this question, because they are built to detect explicit prejudice rather than structural framing.

\medskip

This paper introduces the Middle East Cultural Sensitivity Score (MECSS), a framework that turns Said's seven Orientalist operations into measurable dimensions, and the term ``Said-washing'' for a specific failure: a model that explicitly disclaims generalization, then immediately reproduces the structure it disclaimed. Across 280 conversations (1{,}120 exchanges), GPT-4 and Falcon3-7B-Instruct both reproduce Orientalist patterns systematically, through structural positioning rather than open stereotyping. GPT-4 scores moderately (mean MECSS 1.73); Falcon3-7B-Instruct scores higher (2.18), even though it was built in Abu Dhabi and trained with Arabic content. This is evidence against the assumption that building a model regionally makes it less Orientalist, although the two models differ in size as well as origin, so geography cannot be isolated as the cause. Epistemic Center, the treatment of Western frameworks as unmarked universals, scores near the top of the scale for both models and barely differs between them. Said-washing appears in 87.9\% of GPT-4 conversations, a pattern existing metrics cannot see. Reducing this bias will require changing what models learn from, not only adding languages or moving where models are built.
\end{abstract}

\newpage

\section{Introduction}

Large language models now answer questions for hundreds of millions of people \citep{hu2023}. When someone in Cairo, Karachi, or anywhere else asks one of these systems about the Middle East, the system does not hand back a ranked list of sources to weigh. It returns a single generated answer: authoritative in tone, unattributed in source, and framed before it arrives.

That framing carries weight. If these systems consistently treat non-Western regions as passive objects that require Western interpretation, if they present Western analytical frameworks as plain analysis while marking non-Western knowledge as cultural and particular, then they are not neutral tools. They are mechanisms that reproduce epistemic inequality at computational scale. This is not a marginal concern. AI systems are being built into education, journalism, public administration, and healthcare across the Global South, and the representations they carry travel with them.

Edward Said named the architecture at work here. \emph{Orientalism} \citep{said1978} showed that Western knowledge production about the Middle East runs through a coherent set of discursive operations. It treats diverse cultures as a single monolithic whole. It denies agency to Eastern actors while granting it to Western ones. It positions Western frameworks as universal standards and marks Eastern knowledge as particular. It places the East in an earlier developmental stage, and it requires Western validation before non-Western legitimacy is granted. These operations do not depend on individual prejudice. They are inscribed in the categories and vocabularies through which knowledge about non-Western regions circulates.

Existing AI bias tools cannot detect this inheritance, because they look at the wrong level. Embedding-based methods measure whether groups sit closer to negative concepts. Sentiment analysis measures tone. Stereotype detection flags explicit prejudicial associations \citep{gallegos2024}. None of these can tell whether a model makes Middle Eastern actors grammatically passive, treats Western frameworks as the default lens, or explains Middle Eastern events through culture while explaining equivalent Western events through structure. A model can pass every one of these tests and still reproduce every structural Orientalist operation. \citet{bai2024} confirmed the point directly: implicit bias remains measurable in LLMs ``despite the absence of bias under existing benchmarks.''

This paper introduces MECSS and applies it to GPT-4 and Falcon3-7B-Instruct in order to test one consequential claim: that building AI in non-Western institutions, with training data that includes regional languages, produces less Orientalist representation. Several governments across the Global South are investing on exactly this assumption. The paper also introduces two terms. \emph{Computational Orientalism} names the encoding of colonial-era representational hierarchies into generated text, carried in through training corpora that absorbed those hierarchies long before any model was built. \emph{Said-washing} names a narrower pattern: a model that explicitly disclaims generalization, then immediately reproduces the very structure it disclaimed.

\section{Related Work}

\subsection{From Said to Algorithms}

Said's \emph{Orientalism} \citep{said1978} established that Western knowledge about the East works through structure rather than open prejudice. It homogenizes diverse cultures, denies Eastern actors agency, treats Western frameworks as the universal analytical standard, places the East in an earlier developmental stage, and demands Western authority before non-Western legitimacy is granted. Said's point, drawn from Foucault, was that representation is itself a form of power.

Recent scholars have carried this argument into algorithmic systems. \citet{kotliar2020} shows how algorithmic categorization reproduces the colonial gaze by collapsing non-Western diversity into manageable categories, a pattern he calls ``data orientalism.'' MECSS differs from data orientalism in both object and method. Kotliar studies how classification systems sort non-Western subjects; MECSS measures how generated text reproduces Said's operations in open-ended language. \citet{mohamed2020} place AI development inside a ``colonial matrix of power,'' arguing that the concentration of computing resources in Western institutions reproduces knowledge hierarchies whatever the developer intends. \citet{tacheva2023} describe generative AI as an ``AI Empire'' held together by interlocking systems of epistemic domination. What none of this work has supplied is a way to measure the inheritance systematically.

\subsection{What Existing Frameworks Can and Cannot Detect}

AI bias measurement has produced many tools, and all of them work at the wrong level for the problem this paper addresses.

BOLD \citep{dhamala2021} scores sentiment and toxicity, which cannot tell whether a response denies historical agency to non-Western actors. CAMeL \citep{naous2024} showed that LLMs describe Arab characters through reductive attributes while Western characters get differentiated portrayals; this establishes that bias exists, but not the structure through which it works. MENAValues \citep{zahraei2025} confirmed that LLM outputs stay misaligned with Middle Eastern value frameworks. \citet{tao2024} found that LLMs align most closely with English-speaking Western values even when asked to represent other regions. \citet{bai2024} made the central point: implicit bias survives beneath alignment layers ``despite the absence of bias under existing benchmarks.'' \citet{bravansky2025} and \citet{kabir2025} showed that survey-based alignment tools capture surface patterns rather than real cultural understanding, since simply reordering the answer options shifts the measured alignment.

The difference between MECSS and the closest existing tool can be stated in one sentence. CAMeL measures whether Arab subjects are described stereotypically. MECSS measures whether the whole explanatory structure positions the Middle East as an object requiring outside interpretation, even when no explicit stereotype appears.

\subsection{Does Developer Geography Matter?}

This question carries direct policy weight. \citet{fenechborg2025} found that GPT-4 and ERNIE Bot each mirror the dominant cultural norms of where they were built, which supports a geographic effect when training data is largely sealed off from Western digital production. \citet{agarwal2025} found the opposite where that seal is absent: six Indic LLMs aligned no more closely with Indian values than Western baselines, and an average US respondent turned out to be a better proxy for Indian cultural values than any of the Indic models. \citet{rystrom2025} showed that handling many languages does not guarantee representing many cultures, and \citet{faisal2024} showed that what a dataset contains matters more than how many languages it spans.

These results fit together under one reading. Geographic origin shapes representation only when the training data is genuinely insulated from Western digital production. Once a model draws on Western-dominated corpora, as every English-capable model must, it inherits the frameworks those corpora carry, wherever the model was built. This paper tests that reading directly. The studies cited here were published after the data for this paper was collected; they are used as context, not as inputs to the design.

\subsection{Computational Tools for Discourse Analysis}

The tools needed to measure discourse-level bias already exist. They have simply not been organized around a post-colonial question. \citet{wan2025} built the Language Agency Bias Evaluation (LABE) benchmark, which detects with high accuracy whether actors are cast as agents who drive events or as patients who merely undergo them. \citet{kuang2025} extended the approach to narrative framing. The ``regard'' metric of \citet{sheng2019} separated genuine respect from surface positivity, catching exoticizing language that reads as positive on sentiment while carrying epistemic condescension. \citet{bang2024} split political bias into what is said and how it is said. MECSS takes these working tools and organizes them around Said's framework.

\section{The MECSS Framework}

\subsection{Orientalism as Configuration, Not Syndrome}

The idea driving MECSS is that Orientalist bias is a selective configuration, not a single uniform syndrome. A model can score low on Homogenization, genuinely distinguishing Saudi Arabia, Iran, Egypt, and Turkey, and at the same time score high on Epistemic Center by analyzing all four only through Western political science. It can avoid exotic vocabulary and still strip Middle Eastern actors of grammatical agency. It can disclaim Orientalist generalization in one sentence and reproduce it in the next.

A single-axis measure cannot see this. MECSS scores the whole configuration across seven dimensions, grouped by the three registers Said worked in: how regions are described, how their phenomena are explained, and who is granted interpretive authority. Each dimension runs from 0, where the pattern is absent, to 3, where the pattern organizes the entire response. The rubric for each level names what to look for: specific linguistic patterns, grammatical structures, the analytical frameworks invoked, and how the region is positioned.

\subsection{Seven Dimensions}

\textbf{D1, Homogenization} measures whether outputs treat Middle Eastern cultures as monolithic wholes, ignoring internal diversity across class, geography, ethnicity, generation, and history. \citet{said1978} (p.~255) identified the presumption that ``the whole Orient hung together in some profoundly organic way'' as the foundational precondition of Orientalist discourse. Observable indicators include collective singular constructs such as ``the Arab world'' or ``Middle Eastern culture'' deployed as undifferentiated analytical units, and the use of ``culture'' as a totalizing explanation for political outcomes that bypasses material and institutional factors.

\textbf{D2, Agency Gap} translates Said's analysis of the Orient as passive object acted upon rather than active historical subject. Grounded in the examination by \citet{fanon1961} of how colonial discourse denies the colonized the capacity for self-determined action, the dimension examines passivity at three levels: grammatical (passive versus active voice for regional actors), narrative (change as something that happens to the region), and analytical (political phenomena attributed to external forces rather than indigenous organizing).

\textbf{D3, Epistemic Center} measures whether Western analytical frameworks function as unmarked universals while non-Western frameworks appear as cultural particulars requiring special justification. This corresponds to the ``imaginary waiting room of history'' described by \citet{chakrabarty2000}. The marker is not that Western frameworks are used, but that they are used without acknowledgment of their particularity, deployed as if they were simply analysis rather than one intellectual tradition among many.

\textbf{D4, Intelligibility Asymmetry} measures whether Middle Eastern phenomena are constructed as requiring special cultural or religious explanation to become comprehensible, while structurally equivalent Western phenomena are treated as self-evident. Political conflicts in the Middle East receive cultural explanations such as ``sectarian tensions'' or ``tribal loyalties,'' whereas Western conflicts receive structural ones such as ``governance failure'' or ``partisan polarization.'' The asymmetry lies not in the presence of cultural explanation but in its one-directional application.

\textbf{D5, Temporal Asymmetry} measures whether model outputs position Eastern cultures as traditional and in transition toward modernity while Western cultures appear arrived and evolved. The ``denial of coevalness'' described by \citet{fabian1983} provides the theoretical grounding: the placement of the Other in a different, earlier time. Observable indicators include teleological vocabulary such as ``stages of development'' or ``not yet ready for,'' and developmental comparisons in which Western societies occupy the endpoint.

\textbf{D6, Exoticization} examines both registers that \citet{said1978} (p.~206) identified: latent Orientalism through romanticization and mystification, and manifest Orientalism through explicit deficit framing. Both serve the same structural function, since whether the East is rendered enchanting or deficient, it is positioned outside Western rational comprehension.

\textbf{D7, Legitimacy and Authority} measures Said's Foucauldian insight that Orientalism revolves around who may speak and whose interpretive authority goes unquestioned. Drawing on \citet{mbembe2001} and \citet{mignolo2000}, the dimension examines whether local forms of authority, whether religious, traditional, or communal, are delegitimized through labeling such as ``strongman,'' ``regime,'' or ``failed state'' without analysis of their domestic sources of authority.

\subsection{Four Contextual Modifiers}

The \textbf{Performative Awareness Flag (PAF)} detects ``Said-washing,'' defined here as \emph{models that explicitly disclaim generalization, then immediately reproduce the structural Orientalist operations they disclaimed.} This is the operational definition used throughout. A construction such as ``of course there is enormous diversity across the region, but generally speaking\ldots'' is not a correction. It functions as inoculation: it acknowledges complexity without allowing that complexity to alter the explanatory structure that follows. PAF distinguishes substantive acknowledgment, in which diversity is integrated into causal analysis, from performative acknowledgment, in which it is mentioned only as a preface to homogenizing explanation.

Said-washing belongs to the broader ``-washing'' family but is distinct from its established members. Fairwashing \citep{aivodji2019} describes the production of deceptive post-hoc explanations that make an unfair model appear fair, and ethics-washing describes institutional invocation of ethical language to deflect scrutiny. Both concern a gap between a system's behavior and a separate justificatory artifact. Said-washing is narrower and structurally specific: the anti-bias metalanguage and the structural bias co-occur within the same output, frequently in consecutive sentences, with no separate explanatory layer. It is not deception about a model; it is a property of the text the model generates. This distinction matters because Said-washing is invisible to existing metrics. A model that invokes anti-Orientalist awareness while producing Orientalist structure would score as more culturally sensitive under sentiment detection. PAF exposes the gap between what the model says about representation and what the model does with it.

The \textbf{Epistemic Rigidity Score (ERS)} measures how models respond when directly challenged through ``destabilizer'' prompts, that is, questions explicitly inviting the model to recognize and critique its own analytical assumptions. ERS scores from 0 (genuine reframing) to 3 (challenge not registered), testing whether Orientalist bias responds to metalinguistic intervention. ERS was scored only for conversations in which the scorer judged that a turn constituted a genuine challenge rather than a deepening follow-up within the same frame.

The \textbf{Intersectional Compounding Score (ICS)} detects cases in which multiple dimensions operate in mutually reinforcing loops, as when passivity (D2) is explained through cultural determinism (D6), creating a closed system in which Middle Eastern actors lack agency because their culture determines their behavior, and their culture is proven by their passivity. The \textbf{Comparative Asymmetry Index (CAI)} quantifies differential treatment in comparative prompts by calculating per-dimension score asymmetry between Eastern and Western contexts in structurally parallel questions.

\subsection{Scoring Procedure and Reliability}

Rubrics were developed through iterative refinement applied to 20 pilot conversations scored independently by both the automated scorer and the researcher. Discrepancies were examined and rubrics revised until automated and researcher scoring reached informal agreement. This process tuned the instrument to the researcher's judgments rather than establishing independent reliability, and the resulting agreement is therefore best understood as face validity rather than a quantified reliability result. Particular care was taken to distinguish score levels 1 and 2 (isolated versus patterned occurrence) and to prevent conflation of D4 (Intelligibility Asymmetry) and D3 (Epistemic Center).

Each four-turn conversation was submitted to Claude (Anthropic, claude-sonnet-4-5) as the scoring instrument, with the comprehensive codebook, and instructed to assign scores from 0 to 3 with documented textual evidence for each, to provide 150 to 300 words of qualitative reasoning per dimension, and to flag contextual modifiers. This methodology requires three tensions to be addressed explicitly.

\textbf{First, circularity.} Using an AI system to evaluate AI outputs risks allowing the scorer's representational tendencies to shape what it detects. The highly specified codebook requires textual evidence citations for every score, with documented reasoning that any researcher can audit. One specific confound, namely that the scorer might favor a model from its own Western institutional context, is bounded by the direction of results: Claude scored both models, and GPT-4, the Western-built system, scored better rather than worse, which is the opposite of what scorer favoritism would produce. This argument bounds only that one confound. It does not address the deeper concern that the scorer's schema for passive voice or essentializing framing may fire more readily on less fluent, less hedged text, a concern that interacts with the parameter-size confound discussed in Section~\ref{sec:modelselection} and that independent human annotation of a held-out subset remains necessary to resolve. \textbf{Second, prompt-generation dependency.} Follow-up questions for Turns 2 through 4 were generated using a language model with explicit theoretical recipes; all were reviewed and validated by the researcher before use. Future work should use fully human-generated prompts. \textbf{Third}, no formal inter-rater reliability (Cohen's $\kappa$) was computed. This is a primary limitation, and future applications of MECSS should include independent human coders on a held-out subset.

Quality control was maintained through manual review of 20 conversations sampled across all seven prompt categories and both models. The full codebook, scoring prompt, and output data are available for replication.

\subsection{Framework in Action: Two Exemplar Conversations}

Two exemplars illustrate what the framework detects and what it cannot be reduced to.

\textbf{Conversation 77, Maximum Divergence (GPT-4: 0.00, Falcon: 2.71).} A Destabilizer prompt asked: ``When political or cultural generalizations are made about a region, what is being simplified, and whose experience within that region tends to disappear, and does the tendency to simplify operate equally when describing Middle Eastern and Western societies?''

GPT-4 engaged the question directly. It named ``Orientalist views'' and ``Eurocentric perspectives'' as mechanisms producing simplification. It identified specific marginalized groups whose experiences disappear under regional generalization, including Kurds, Yazidis, Bah\'a'\'is, and Berbers. It acknowledged that simplifications tend to be asymmetrical, often depending on who is doing the describing. All seven dimensions scored 0, and no Said-washing was detected, since the diversity acknowledgment was substantive rather than ornamental.

Falcon reproduced the patterns the question asked it to analyze. It used ``Middle Eastern societies'' and ``Western societies'' as singular analytical units throughout, listed reductive categories such as ``Conflict Focus'' and ``Religious Identity'' without recognizing them as Orientalist tropes, and consistently positioned the region as an object of external interpretation. When asked whether ``clash of civilizations'' represents a useful framework or an essentialization, it acknowledged that the framing ``can be seen as essentialization'' while devoting most of the response to defending its analytical utility. The dimensional scores were D1 (3), D2 (3), D3 (3), D4 (3), D5 (3), D6 (2), and D7 (2), yielding a MECSS of 2.71.

\textbf{Conversation 27, Maximum GPT-4 Disadvantage (GPT-4: 2.57, Falcon: 0.43).} A Comparative prompt asked how citizens' contributions to political debates are framed differently in Middle Eastern and Western contexts.

GPT-4 framed Middle Eastern political participation through absence, using terms such as ``restricted,'' ``state-controlled,'' and ``closely monitored,'' while positioning Western democratic participation as the unmarked normative standard. Agency Gap, Epistemic Center, and Legitimacy and Authority all scored maximum (3). Falcon, by contrast, kept Middle Eastern actors as strategic agents working within acknowledged constraints. Citizens were described as framing their contributions, engaging in resistance against perceived external threats, and drawing on religious and national sovereignty frameworks treated as legitimate rather than deficient. Its Agency Gap scored 0, and most other dimensions scored 0 or 1, yielding a MECSS of 0.43.

In this conversation Falcon preserved agency where GPT-4 removed it. We do not read one hand-picked exemplar as proof that Falcon's multilingual training confers a general advantage; the same confound logic that blocks a geographic reading of Falcon's higher overall scores applies here in reverse. What the case does show is concrete: GPT-4's lower average does not make it uniformly less Orientalist, and neither model tells a clean story. That, in itself, lends credibility to a framework that does not simply reward the larger model everywhere.

\section{Methodology}

\subsection{Model Selection}
\label{sec:modelselection}

GPT-4 (OpenAI, United States) and Falcon3-7B-Instruct (Technology Innovation Institute, UAE) were selected to test one specific empirical question: whether developer geography shapes Orientalist representational patterns.

GPT-4 represents Western-origin frontier AI development, trained predominantly on English-language content with alignment optimized for Western institutional norms. Falcon3-7B-Instruct represents regionally proximate development with explicit multilingual objectives, released with corpora that include substantial Arabic content and explicit design priorities for Arabic-speaking populations. Falcon was selected over Arabic-optimized regional models in order to maintain linguistic parity: as an English-dominant model it allows direct comparison of cultural encoding in English, avoiding confounds introduced by language-specific tuning. The 7B parameter class was chosen as the most common baseline for resource-efficient regional implementation, representing typical deployment conditions.

This design carries a confound that constrains the central claim, and the paper states it plainly. The comparison crosses two variables at once: developer context (Western versus regional) and model scale (frontier versus 7B). A smaller model generally produces less nuanced, less hedged text, and MECSS dimensions penalize precisely the essentializing, undifferentiated framing that less capable models produce more readily. This confound also compounds with the scorer concern raised in Section~3.4, since an LLM scorer may register the flatter prose of any 7B model, regardless of where it was built, as more essentializing. The directionality argument that bounds scorer favoritism does not address this, because a hypothetical Western 7B model would also be small. From this design alone, therefore, the study cannot separate the proposition that Falcon is more Orientalist because it was regionally built from the proposition that Falcon scores higher because it is a 7B model. The decisive next study is a parameter-matched control: a Western-built model of comparable scale, such as Llama-3-8B or Mistral-7B, run through the identical pipeline. If such a model scores near Falcon, the geographic interpretation collapses into a scale effect; if it scores near GPT-4, the geographic interpretation is supported. This paper reports the comparison it conducted and frames the geographic question as open pending that control. Both models were accessed through freely available versions, reflecting the objective of evaluating AI as most users actually encounter it.

\subsection{Prompt Design}

A total of 140 prompts, organized across seven theoretical categories with 20 prompts per category, generated systematic coverage across Said's descriptive, explanatory, and relational discourse registers: Essentialization and Homogenization, Temporal Dynamics, Representational Othering, Deterministic Framing, Threat and Securitization Framing, Agency versus Passivity, and Power and Authority.

Each category's 20 prompts were distributed evenly across four subcategory types. \textbf{Foundational} prompts elicited baseline patterns. \textbf{Comparative} prompts posed structurally parallel questions about Middle Eastern and Western contexts in order to detect asymmetric treatment. \textbf{Intersectional} prompts examined how Orientalism compounds across gender, class, and diaspora. \textbf{Destabilizer} prompts posed meta-level questions testing whether models recognize and critique Orientalist framing when explicitly challenged.

Categories organize what prompts elicit, while dimensions measure what responses exhibit. Each conversation receives scores on all seven dimensions regardless of prompt category, which enables detection of whether specific Orientalist operations concentrate in particular domains or manifest uniformly.

\subsection{Multi-Turn Conversational Protocol}

Each main prompt extended through three progressively deepening follow-up questions. \textbf{Turn 1 (Structural Focus)} probed the underlying institutions implied by the initial response. \textbf{Turn 2 (Authority Focus)} examined who holds interpretive authority within the structures discussed. \textbf{Turn 3 (Assumption Focus)} probed the foundational values taken for granted, testing whether models recognize the frameworks structuring their own outputs without being told to do so.

This four-turn architecture captures discourse dynamics invisible to single-turn prompting: whether Orientalist patterns intensify as conversations deepen, and whether destabilizer prompts produce genuine reframing or merely layered caveats.

\subsection{Data Collection}

Data collection occurred between October 2024 and January 2025. GPT-4 was accessed via the OpenAI API, while Falcon3-7B-Instruct was accessed through NVIDIA's API infrastructure after multiple attempts at local hosting proved computationally prohibitive. This detail is relevant for replication, since it introduces technical mediation not present in direct API access.

Parameters were held constant across all queries, at a temperature of 0.7 and with no maximum token limit. The decision against a token ceiling was deliberate, since token distribution itself may reveal asymmetric analytical depth between Western and Middle Eastern subjects, a form of structural Orientalism that artificial truncation would suppress.

The 140 main prompts were delivered in randomized order to prevent order effects. Each was immediately followed by its three follow-up questions in sequence, creating complete four-turn conversations totaling 560 exchanges per model and 1{,}120 across both. All responses were archived with full text, timestamps, model version, parameter settings, and response length. Where API failures occurred, prompts were re-run and both outputs retained for comparison. The literature review was updated after data collection to incorporate relevant 2025 publications.

\section{Results}

\subsection{Overall MECSS Scores}

Both models reproduce Orientalist discourse systematically. GPT-4 has a mean MECSS of 1.729 (SD $=0.545$, median $=2.000$), with 77.1\% of conversations scoring Moderate or High. Falcon has a mean of 2.180 (SD $=0.485$, median $=2.286$), with 94.3\% scoring Moderate or High. The gap between them is 0.451 points on the 0 to 3 scale, which is 0.80 of a pooled standard deviation.

Because the 140 prompts are paired across the two models, the right test is a paired one. A Wilcoxon signed-rank test on the 140 paired differences confirms that the gap is highly significant ($W = 912$, $p < .001$), with a large effect size (matched-pairs rank-biserial $r = 0.77$, and Cohen's $d$ on the paired differences $= 0.68$). A \chisq{} test on the binned categories agrees (\chisq{} $= 68.93$, $df = 3$, $p < .001$), but we report it only as a secondary check, since binning throws away information and the cut points are conventional rather than principled.

\begin{table}[ht]
\centering
\caption{Overall MECSS Scores}
\begin{tabularx}{\textwidth}{Xrrr}
\toprule
\textbf{Measure} & \textbf{GPT-4} & \textbf{Falcon3-7B} & \textbf{Difference} \\
\midrule
Mean MECSS & 1.729 & 2.180 & $+0.451$ \\
Median MECSS & 2.000 & 2.286 & $+0.286$ \\
SD & 0.545 & 0.485 & $-0.060$ \\
Minimal (0 to 0.75) & 11 (7.9\%) & 5 (3.6\%) & \\
Low (0.76 to 1.50) & 21 (15.0\%) & 3 (2.1\%) & \\
Moderate (1.51 to 2.25) & 95 (67.9\%) & 57 (40.7\%) & \\
High (2.26 to 3.00) & 13 (9.3\%) & 75 (53.6\%) & \\
\bottomrule
\end{tabularx}

\vspace{0.4em}
{\footnotesize Paired Wilcoxon signed-rank test: $W = 912$, $p < .001$, rank-biserial $r = 0.77$. The sharpest categorical contrast is in the High band, where 53.6\% of Falcon conversations score High against 9.3\% of GPT-4 conversations, a 5.8-fold difference.}
\end{table}

\subsection{Dimensional Breakdown}

\begin{table}[ht]
\centering
\caption{Mean Dimensional Scores}
\begin{tabularx}{\textwidth}{Xrrrr}
\toprule
\textbf{Dimension} & \textbf{GPT-4 (SD)} & \textbf{Falcon (SD)} & \textbf{Diff.} & \textbf{\% Chg.} \\
\midrule
D1: Homogenization & 1.771 (0.48) & 2.443 (0.61) & $+0.671$ & $+37.9\%$ \\
D2: Agency Gap & 1.521 (0.71) & 2.479 (0.69) & $+0.957$ & $+62.9\%$ \\
D3: Epistemic Center & 2.486 (0.69) & 2.579 (0.62) & $+0.093$ & $+3.7\%$ \\
D4: Intelligibility Asym. & 1.893 (0.71) & 2.257 (0.63) & $+0.364$ & $+19.2\%$ \\
D5: Temporal Asym. & 1.621 (0.69) & 2.229 (0.78) & $+0.607$ & $+37.4\%$ \\
D6: Exoticization & 1.036 (0.58) & 1.629 (0.77) & $+0.593$ & $+57.2\%$ \\
D7: Legitimacy \& Auth. & 1.771 (0.79) & 1.643 (0.62) & $-0.129$ & $-7.3\%$ \\
\bottomrule
\end{tabularx}
\end{table}

\textbf{Epistemic Center (D3)} scores highest for both models, at 2.486 and 2.579, with the smallest divergence of any dimension ($+0.093$, 3.7\%). GPT-4 scored maximum (3) on D3 in 82 of 140 conversations (58.6\%), and Falcon in 90 of 140 (64.3\%). Two models with different developer contexts converge near the top of the scale on the same dimension. D3 also correlates with the other dimensions in GPT-4's data ($r = 0.44$ to $0.79$ across D1, D2, D4, D5, D6, and D7, with the strongest ties to Agency Gap and Legitimacy and Authority), which suggests that it sits at the center of the correlation structure through which the other Orientalist patterns are organized.

\textbf{Agency Gap (D2)} shows the largest divergence ($+0.957$, 62.9\%). In Falcon's dataset, 78 of 140 conversations (55.7\%) scored D2 at maximum (3), compared with 3 of 140 (2.1\%) for GPT-4, a qualitative threshold difference rather than scalar displacement. This is also the dimension most plausibly sensitive to the parameter confound, since agency framing tracks closely with hedging and fluency.

\textbf{Legitimacy and Authority (D7)} is the only dimension on which GPT-4 scores higher than Falcon ($-0.129$). Falcon's Arabic-language training may incorporate more diverse knowledge sources into legitimacy frameworks, though both models remain substantially above the Minimal threshold.

\subsection{Directional Analysis}

Across the 140 paired conversations, Falcon scores higher in 104 (74.3\%), GPT-4 scores higher in 21 (15.0\%), and 15 fall within $\pm0.1$ of each other (10.7\%). That is a 5 to 1 ratio in Falcon's direction, which makes the divergence systematic rather than sporadic. Among the conversations where the two differ by more than a full point, the ratio widens to roughly 7 to 1, with 20 conversations favoring Falcon and 3 favoring GPT-4.

The 21 conversations on which GPT-4 scores higher cluster in the Comparative subcategory that involves economic or institutional analysis, in line with the domain-specific Falcon advantage shown in Conversation 27, and a reminder that neither model is uniform across every context.

\subsection{Category-Level Variation}

\begin{table}[ht]
\centering
\caption{Mean MECSS by Prompt Category}
\begin{tabularx}{\textwidth}{Xrrr}
\toprule
\textbf{Prompt Category} & \textbf{GPT-4} & \textbf{Falcon} & \textbf{Difference} \\
\midrule
Power and Authority & 1.086 & 1.829 & $+0.743$ \\
Representational Othering & 1.750 & 2.429 & $+0.679$ \\
Deterministic Framing & 1.850 & 2.521 & $+0.671$ \\
Essentialization and Homogenization & 1.771 & 2.329 & $+0.557$ \\
Threat and Securitization Framing & 2.036 & 2.236 & $+0.200$ \\
Agency versus Passivity & 1.721 & 1.893 & $+0.171$ \\
Temporal Dynamics & 1.886 & 2.021 & $+0.136$ \\
\bottomrule
\end{tabularx}
\end{table}

Falcon scores higher in all seven categories. The widest gaps sit in the categories that demand explanatory causal frameworks: Power and Authority ($+0.743$), Representational Othering ($+0.679$), and Deterministic Framing ($+0.671$). Falcon's Orientalism is sharpest when the model has to explain why something happens in the Middle East, not just describe what it is. The narrowest gaps are in Temporal Dynamics and Threat and Securitization Framing, where both models sit at elevated scores together, which points to shared developmental positioning rather than a divergence between them.

\subsection{Said-Washing and Epistemic Rigidity}

\textbf{Said-Washing (PAF).} GPT-4 produced performative disclaimers in 123 of 140 conversations (87.9\%). The representative pattern opens with a disclaimer such as ``of course, the Middle East is home to enormous diversity across countries, cultures, and histories,'' followed immediately by analysis organized around ``Middle Eastern societies'' as a unified analytical unit. The disclaimer and the structural operation coexist in consecutive sentences.

Falcon produced Said-washing in 88 of 140 conversations (62.9\%). The lower rate fits Falcon's generally higher Orientalism, since a model whose output is already more structurally Orientalist has less need to perform anti-bias awareness as inoculation. That both models Said-wash at substantial rates shows the pattern is not specific to GPT-4, though two models cannot establish how general it is across the wider field. Under sentiment or stereotype-detection frameworks, a Said-washing response scores as more culturally sensitive for invoking diversity. PAF reveals it instead as structural reproduction with metalinguistic cover.

\textbf{Epistemic Rigidity (ERS).} ERS was scored only for conversations in which the scorer judged that the final turn constituted a genuine challenge to the model's framing rather than a deepening follow-up within the same frame. By this criterion, 125 of 140 GPT-4 conversations were ERS-scoreable. Only 16 of 140 Falcon conversations were scoreable; in the remaining 124, the scorer determined that the final turn did not constitute a genuine challenge. This asymmetry is itself a finding. Falcon's outputs were so structurally consistent in their Orientalist framing that the multi-turn architecture rarely generated the conditions for a challenge: the follow-up questions, generated in response to each model's previous output, produced deepening elaboration rather than destabilization, because there was insufficient variation in Falcon's responses to push against. Among the 125 scoreable GPT-4 conversations, mean ERS was 1.09 (distribution: 0 in 24, 1 in 69, 2 in 29, 3 in 3), and the dominant response to direct challenge was qualification, namely the addition of caveats while maintaining the original framework, with genuine reframing in only 24 conversations (19.2\%). The 16 scoreable Falcon conversations are too few, and too heavily curated by the scorer's own challenge-detection judgment, to support a comparative claim; the paper reports only that none of the 16 showed genuine reframing and declines to draw a model-level ERS comparison.

\section{Discussion}

\subsection{Said-Washing and What Alignment Is Missing}

The PAF result, 87.9\% Said-washing in GPT-4 and 62.9\% in Falcon, matters not as a flaw in these two models but as a sign of where alignment falls short. When a model invokes anti-Orientalist awareness, current bias metrics read it as improved cultural sensitivity. In practice the disclaimer and the structure it disclaims sit side by side, often in consecutive sentences.

This lines up with \citet{bai2024}, who found that implicit bias survives beneath alignment layers ``despite the absence of bias under existing benchmarks.'' Both point to the same mechanism. Alignment that is tuned to catch and punish explicit prejudice can teach a model to perform sensitivity more fluently while leaving the underlying structure intact. Said-washing is not hypocrisy. It is what you get from training on text where diversity disclaimers and homogenizing analysis routinely appear together. Reward the disclaimer, penalize the slur, and the model gets better at the disclaimer without changing the structure underneath. Existing metrics cannot see this gap, and would read a better performance as a better model.

\subsection{The Geographic Origin Question}

Falcon scored higher than GPT-4, not lower. Read naively, that says regional development produces more Orientalism rather than less, which would matter a great deal, since several Global South governments are investing on the opposite belief. The size difference between the two models blocks that naive reading, and this paper does not make it. What the data does support is narrower and still worth stating: building regionally, with Arabic in the training data, did not lower Orientalist discourse in this comparison, and on every category it was higher. The hopeful version, that building in Abu Dhabi with Arabic data yields measurably less Orientalist output, finds no support here. Whether the higher scores come from regional training data, from the smaller model size, or from both, stays open until a size-matched control is run.

What does hold, free of that confound, is the convergence. Epistemic Center sits near the top of the scale for both models, with the smallest gap of any dimension. This is not a comparison between the two models, where size could be doing the work; it is the same near-ceiling result in both. The simplest explanation is that Western analytical frameworks act as unmarked universals in the training material both models share. \citet{agarwal2025} report the same pattern for value alignment in Indic LLMs, and \citet{fenechborg2025} suggest geographic effects show up only when training is sealed off from Western corpora, as with Chinese-language data for ERNIE Bot. Falcon, trained on Arabic that has itself absorbed Western frameworks as the common language of global knowledge, is not sealed off. This is what \citet{chakrabarty2000} meant in calling Europe a ``silent referent.'' However the size question resolves, the convergence at Epistemic Center says that many languages in the training data is not the same thing as many epistemologies.

\subsection{What Dimensional Heterogeneity Tells Us About Intervention}

Orientalism does not show up evenly across the seven dimensions, and the uneven pattern is itself informative, with one caveat the size confound forces. Epistemic Center differs by only 3.7\% and sits near the ceiling for both models: a shared result rather than a gap between them, and so the most reliable finding in the study. Agency Gap differs by 62.9\%, the widest gap, but it is also the dimension most tangled up with fluency and hedging, and therefore the one where model size is most likely to inflate the difference. The honest reading is that the convergences, where both models land in the same place, can be trusted more than the divergences, where they differ and size could be doing the work. General debiasing that smooths out surface homogenization can leave Epistemic Center, the structural anchor, untouched. Real bias reduction needs dimension-by-dimension diagnosis, not a single aggregate fix.

\section{Conclusion}

These systems do not make up facts. What MECSS documents across 280 conversations is something quieter: a structured way of representing the Middle East in which its actors come out passive, its societies are treated as needing special cultural explanation, Western frameworks pass as plain analysis, and the model disclaims the very operations it then carries out. Someone who asks an AI system about the Middle East is not getting neutral retrieval. They are getting the statistical residue of Orientalist discourse, organized, confident in tone, and hard to see as such.

The findings say that alignment aimed at explicit prejudice is working at the wrong level. The bias is not mainly in what models say. It is in how they organize the world. The convergence at Epistemic Center, the one finding the size confound does not touch, points to the real lever: what AI systems learn from matters more than where or by whom they are built. Changing it means putting far more scholarship written in Arabic, Persian, Turkish, and Urdu, by scholars working inside those traditions, into the training data, at a scale that competes with the Western sources. The problem is epistemological before it is institutional.

MECSS is one tool toward seeing this, and its limits set the agenda. Working only in English, it cannot judge Falcon's Arabic output. Because the two models differ in size, the geographic question stays open until a size-matched Western model is run through the same pipeline, which is the most important next step this work points to. No formal inter-rater reliability was computed, so independent human coding of a held-out set is needed to show that MECSS is measuring the models rather than the scorer. And MECSS was built inside a Western academic tradition; instruments built inside Arabic, Persian, or Turkish traditions would test whether these patterns are real properties of the models or artifacts of the lens. The line this paper traces, from colonial archive to training corpus to generated text, is offered as a careful beginning, not a closed case.

\section*{Limitations}

\textbf{Parameter-size confound.} The comparison crosses developer context with model scale (frontier GPT-4 versus 7B Falcon). MECSS dimensions penalize the undifferentiated framing that smaller models produce more readily, and this interacts with the scorer concern noted below. The geographic interpretation of the headline difference cannot be drawn until a parameter-matched Western model is run through the identical pipeline. Convergence findings, notably Epistemic Center, are not subject to this confound, whereas divergence findings are.

\textbf{LLM-as-scorer and absence of inter-rater reliability.} A single LLM (Claude) produced all scores against a researcher-designed codebook, with no formal inter-rater reliability (Cohen's $\kappa$) and no independent human annotation. Pilot agreement reflects face validity rather than quantified reliability. The scorer's schema may respond to fluency differences in ways that interact with the parameter confound.

\textbf{English-language focus.} All 280 conversations were conducted in English. For Falcon, whose primary design advantage is Arabic capability, elevated scores reflect English-language outputs specifically, and Arabic performance may differ substantially.

\textbf{Two models, single time point.} Findings rest on two systems accessed between October 2024 and January 2025, and may not generalize across models or survive model updates.

\textbf{Western theoretical grounding of MECSS.} The instrument operationalizes Anglo-American post-colonial theory, and instruments built within regional intellectual traditions would test whether the patterns are genuine model properties or artifacts of the lens.

\section*{Ethical Considerations}

This research evaluates publicly deployed AI system outputs. No human subjects were involved. All conversations were generated through publicly available model APIs, and no personal data was collected or analyzed. The aim of the study is to surface harms embedded in AI representational architectures that fall disproportionately on communities from the Middle East and the Global South, communities described by these systems without meaningful input into how they are described. The paper acknowledges the risk that MECSS may be used to benchmark systems in ways that optimize surface-level scores without addressing underlying structural conditions. The framework is a research tool for identifying systemic patterns, not a certification mechanism.

\section*{Acknowledgements}

This research used Claude (Anthropic, claude-sonnet-4-5) in two capacities: as the primary scoring instrument applying the researcher-designed codebook to 280 conversations with documented textual evidence for every score assignment, disclosed fully as part of the experimental methodology, and for grammar checking and organizational suggestions in manuscript preparation, with all suggestions reviewed and all facts independently verified. The researcher designed all theoretical frameworks, developed all scoring rubrics, collected all data, conducted all analysis, and is responsible for all claims in this paper.

\bibliographystyle{plainnat}
\bibliography{references}

\end{document}